\documentclass{article}

\usepackage{microtype}
\usepackage{graphicx}
\usepackage{subcaption}
\usepackage{booktabs} 
\usepackage{multirow}
\usepackage{hyperref}
\usepackage{xurl}
\usepackage[table]{xcolor} 
\definecolor{bestbg}{RGB}{198,234,212}   
\definecolor{secondbg}{RGB}{231,242,255} 
\definecolor{mygreen}{HTML}{00AA00}
\newcommand{\best}[1]{\cellcolor{bestbg}\textbf{#1}}
\newcommand{\second}[1]{\cellcolor{secondbg}#1}

\definecolor{gaincolor}{HTML}{D32F2F}  
\definecolor{speedcolor}{HTML}{1976D2} 

\newcommand{\accgain}[1]{%
  \rlap{\kern1pt\raisebox{-0.8ex}{\fontsize{5}{0}\selectfont\textbf{\textcolor{gaincolor}{+#1}}}}%
}

\newcommand{\speedup}[1]{%
  \rlap{\kern1pt\raisebox{-0.8ex}{\fontsize{5}{0}\selectfont\textbf{\textcolor{speedcolor}{#1$\times$}}}}%
}

\usepackage[preprint]{icml2026}

\usepackage{amsmath}
\usepackage{amssymb}
\usepackage{mathtools}
\usepackage{amsthm}

\usepackage[capitalize,noabbrev]{cleveref}

\theoremstyle{plain}

\theoremstyle{definition}

\theoremstyle{remark}

\usepackage[textsize=tiny]{todonotes}
\usepackage{bm}
\icmltitlerunning{LLM Latent Reasoning as Chain of Superposition}

\begin{document}

\twocolumn[
  \icmltitle{LLM Latent Reasoning as Chain of Superposition}

  \icmlsetsymbol{equal}{*}

  \begin{icmlauthorlist}
    \icmlauthor{Jingcheng Deng}{ict,ucas}
    \icmlauthor{Liang Pang}{ict}
    \icmlauthor{Zihao Wei}{ict,ucas}
    \icmlauthor{Shicheng Xu}{ict,ucas}
    \icmlauthor{Zenghao Duan}{ict,ucas}
    \icmlauthor{Kun Xu}{}   
    \icmlauthor{Yang Song}{} 
    \icmlauthor{Huawei Shen}{ict,ucas}
    \icmlauthor{Xueqi Cheng}{ict,ucas}
  \end{icmlauthorlist}

  \icmlaffiliation{ict}{State Key Laboratory of AI Safety, Institute of Computing Technology, Chinese Academy of Sciences, Beijing, China}
  \icmlaffiliation{ucas}{University of Chinese Academy of Sciences, Beijing, China}

  \icmlcorrespondingauthor{Jingcheng Deng}{dengjingcheng23s@ict.ac.cn}
  \icmlcorrespondingauthor{Liang Pang}{pangliang@ict.ac.cn}

  \icmlkeywords{Machine Learning, ICML}

  \vskip 0.3in
]


\printAffiliationsAndNotice{}  

\begin{abstract}
  Latent reasoning offers a computation-efficient alternative to Chain-of-Thought but often suffers from performance degradation due to distributional misalignment and ambiguous chain definitions. Ideally, latent reasoning should function as a superposition of multiple reasoning paths. To realize this, we introduce Latent-SFT, a unified framework addressing challenges at three levels: token, chain, and learning. First, we define the Latent-Vocab to constrain hidden states within the pre-trained vocab-space. Second, we construct the Latent-Chain via Induction-Supervision Masking to ensure semantic compactness and sufficiency. Third, we employ Latent-Optim with stochastic Gumbel-Softmax to guide the model toward generalizable solutions. Empirical results demonstrate that Latent-SFT consistently outperforms explicit SFT across six mathematical benchmarks (e.g., GSM8k, AIME24) while achieving a 2.7x to 5.5x reduction in reasoning length. Analysis confirms that our method effectively captures a superposition of diverse reasoning trajectories rather than merely compressing a single path.
  The models and datasets of our method are available in a repository: \url{https://anonymous.4open.science/r/Latent-SFT-CF17}.
\end{abstract}

\begin{figure}[!htbp]
  \centering
  \includegraphics[width=1.0\linewidth]{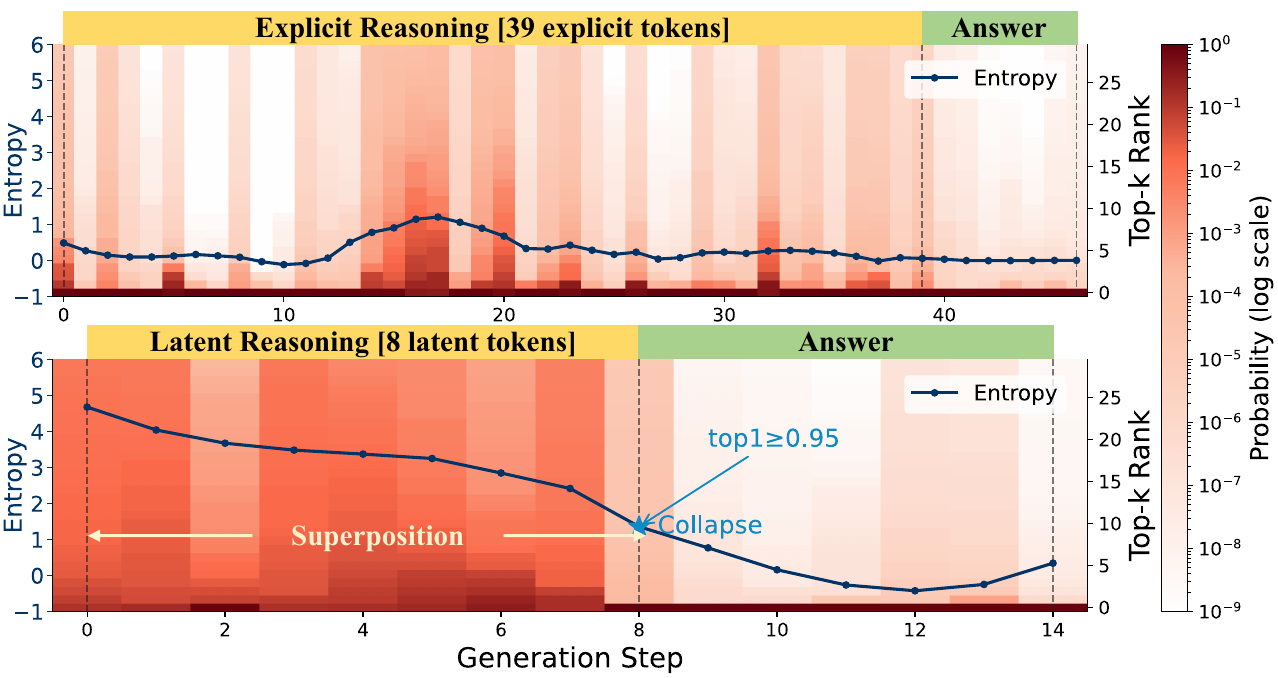}
  \caption{Comparison of Latent and Explicit Reasoning. Latent reasoning operates as a superposition state, where each step carries a higher entropy. Therefore, it can complete the reasoning process and derive the correct answer using fewer tokens.}
  \label{fig_entropy_var}
\end{figure}

\section{Introduction}
Large language models (LLMs) have shown strong reasoning abilities \citep{zhao2024large} across diverse tasks \citep{cao2025pretrainingtestsetlonger,DBLP:journals/corr/abs-2508-20038,DBLP:conf/iclr/DengWPDSC25,DBLP:journals/corr/abs-2502-11401,DBLP:conf/coling/WeiDPDSC25}. A key driver of this success is chain-of-thought prompting \citep{DBLP:conf/nips/Wei0SBIXCLZ22, DBLP:conf/www/XuPSCC24, DBLP:conf/nips/YaoYZS00N23}, which solves complex problems by generating intermediate steps in natural language. However, current reasoning models \citep{qwq32b,DBLP:journals/corr/abs-2501-12948} rely on long chains of tokens \citep{DBLP:journals/corr/abs-2505-16142}, leading to redundancy and high computational cost. Latent Reasoning is a promising paradigm that significantly reduces the length of reasoning chains by shifting computation to a continuous latent space. This paradigm theoretically enhances information bandwidth and exploration efficiency \citep{DBLP:journals/corr/abs-2507-06203}, yet current methods frequently suffer from severe performance degradation compared to explicit reasoning \citep{DBLP:journals/corr/abs-2412-06769}.

Ideally, latent reasoning should function as a superposition of multiple reasoning paths, enabling the model to explore diverse logical trajectories simultaneously within a latent space. However, realizing this potential is impeded by three hierarchical flaws in current methodologies. First, at the \textbf{latent token level}, existing approaches rely on raw, unconstrained hidden states \citep{DBLP:journals/corr/abs-2412-06769,DBLP:journals/corr/abs-2412-13171}. These high-dimensional vectors suffer from severe distributional shifts relative to the model's vocab-space, making them difficult for the model to interpret. Second, at the \textbf{latent chain level}, current definitions are ambiguous; they typically assume the latent chain need only decode the final answer \citep{DBLP:journals/corr/abs-2502-21074}. This "black box" assumption neglects intermediate semantic fidelity, failing to capture the step-by-step logic required for complex deduction. Third, at the \textbf{latent learning level}, optimization is inefficient. Models are updated solely via indirect backpropagation from the final explicit supervision \citep{DBLP:journals/corr/abs-2412-06769}, lacking the step-by-step guidance necessary to learn stable reasoning trajectories.



To systematically address these challenges, we introduce \textbf{Latent-SFT}, a comprehensive framework that unifies token-level representation, chain-level definition, and learning strategies. For token-level representation, we formally define the \textbf{Latent-Vocab} as a superposition within the vocab-space (Section~\ref{sec_preliminary}). This formulation aligns latent tokens with the model's inherent vocabulary distribution. For the chain-level definition, we introduce the \textbf{Latent-Chain} (Section~\ref{sec:latent_chain}), constructed via Induction-Supervision Masking to satisfy three critical semantic properties. First, the Latent Token Induction Mask (LTIM) enforces semantic compactness by compressing explicit reasoning segments into dense representations. Second, these representations are formalized as a Latent-Vocab—a superposition within the vocab-space—ensuring semantic compatibility with the model's semantic manifold. Third, the Latent Token Supervision Mask (LTSuM) guarantees informational sufficiency by mandating that these tokens can reconstruct subsequent reasoning steps. Finally, regarding the learning strategy, to enable the LLM to autonomously generate these latent chains, we propose \textbf{Latent-Optim}. Within this framework, we address the static nature of the constructed latent chains by employing stochastic Gumbel-Softmax optimization. This technique injects Gumbel noise to approximate sampling, acting as an implicit Hessian regularizer (see Theorem 1 in Appendix~\ref{app:proof}) that guides the model toward flatter, more generalizable minima.


As illustrated in Figure~\ref{fig_entropy_var}, our model realizes the ideal latent reasoning via a chain of superposition analogous to a quantum wavefunction. Upon reasoning convergence, the system undergoes entropy collapse, precipitating the latent superposition into a discrete linguistic eigenstate. Empirical results demonstrate that Latent-SFT consistently outperforms explicit SFT models across six diverse mathematical reasoning benchmarks. Notably, it achieves a 2.7x to 5.5x reduction in reasoning chain length while surpassing prior latent reasoning approaches. Further analysis reveals that its latent reasoning is not merely a compression of a single chain, but rather a superposition of multiple reasoning chains.

\section{Related Work}
Current research remains entangled in debates over the reasonable representation of latent tokens, while the critical dimensions of latent chain definition and optimization strategies have been largely overlooked. Consequently, the definition of the latent token serves as the primary differentiator in this field. We categorize existing research based on this pivotal factor.

\begin{figure}[t]
  \centering
  \includegraphics[width=\linewidth]{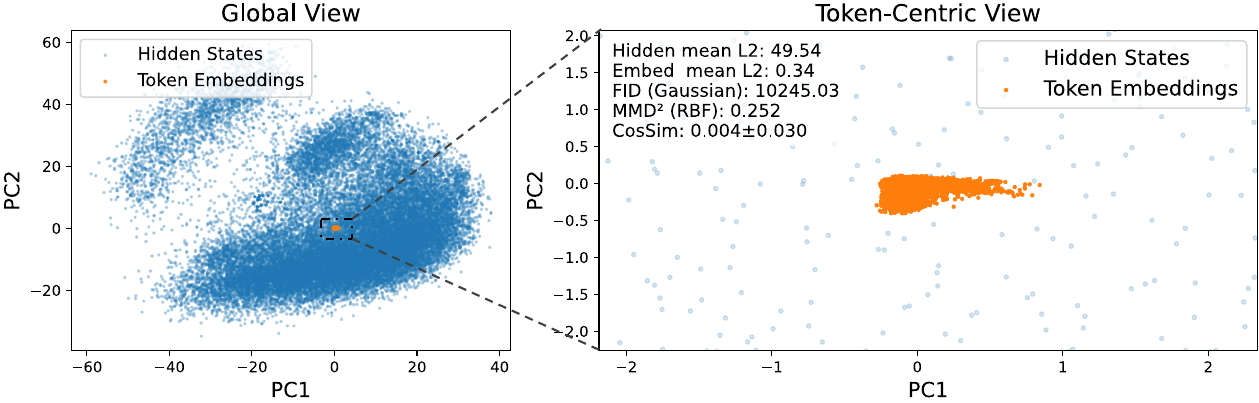}
  \caption{Visualization of last-layer hidden states and token embeddings of the LLaMA-3.2-1B-Instruct model on the GSM8k dataset. Results are shown from both global and token-specific perspectives to highlight the substantial distributional differences. In addition, we report the statistical differences between the two distributions using the Fréchet Inception Distance (FID) and the squared Maximum Mean Discrepancy (MMD$^2$). We also estimate cosine similarity via random sampling.}
  \label{fig_hidden_vs_embeddings}
\end{figure}

\paragraph{Unconstrained Representations.} 
Pioneering approaches, such as COCONUT \citep{DBLP:journals/corr/abs-2412-06769} and CCOT \citep{DBLP:journals/corr/abs-2412-13171}, adopted a direct approach by utilizing the last-layer hidden states as latent tokens. While computationally straightforward, these methods rely on raw, high-dimensional vectors that are unconstrained by the model's vocabulary distribution. Although subsequent works like CODI \citep{DBLP:journals/corr/abs-2502-21074} and PCCOT \citep{DBLP:journals/corr/abs-2506-18582} attempted to mitigate this misalignment via MLP mapping or distillation, they fundamentally remain bound to the continuous hidden state paradigm. Consequently, without strict constraints aligned with the pre-trained semantic manifold, these methods often face optimization instability and limited performance gains compared to explicit reasoning.
\begin{figure}[t]
  \centering
  \includegraphics[width=0.7\linewidth]{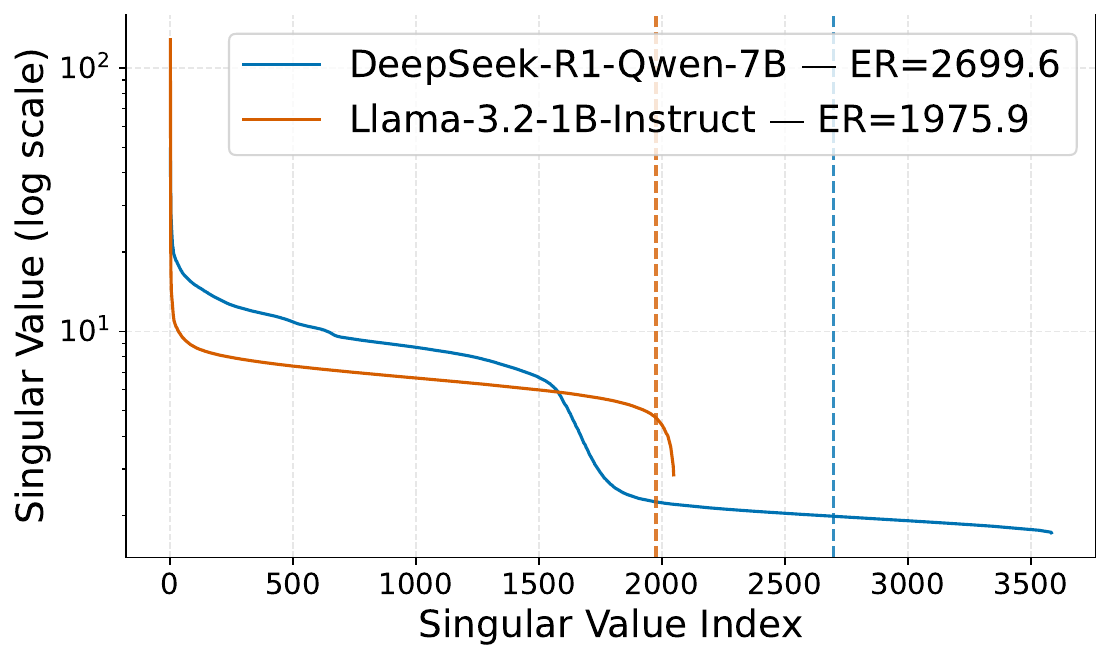}
  \caption{Decay curves of singular values for embedding matrices in various LLMs. The legend indicates the effective rank of each embedding matrix.}
  \label{fig_svd_decay}
  \vspace{-0.3cm}
\end{figure}

\paragraph{Constrained Representations.}
To address the distributional gap, recent studies have introduced structural constraints. CoLaR \citep{DBLP:journals/corr/abs-2505-16552} represents a step forward by modeling latent tokens via Gaussian distributions, achieving improved alignment through a dedicated latent head. Similarly, methods like Soft Thinking \citep{DBLP:journals/corr/abs-2505-15778} and Mixture-of-Input (MoI) \citep{DBLP:journals/corr/abs-2505-14827} attempt to bridge the gap by constructing latent tokens as linear combinations of vocabulary embeddings. However, the performance gains of these embedding-merge approaches remain marginal, and crucially, they fail to reduce the reasoning chain length. As analyzed by \cite{wu2025llmssinglethreadedreasonersdemystifying}, these training-free methods operate essentially as greedy decoding in a latent space. Consequently, they are incapable of realizing the ideal state of latent reasoning—namely, the learned superposition of multiple reasoning trajectories. In contrast, we formally define the Latent-Vocab and Latent-Chain to enable simultaneous multi-path exploration, thereby significantly shortening reasoning chains while surpassing existing baselines. Given that training-free methods neither involve parameter optimization nor achieve the reasoning length reduction central to our contribution, we exclude them from our primary baselines to focus on training-based methods that deliver compression of the reasoning chain.

\section{Latent-Vocab:  Superposition of Vocab-Space}
\label{sec_preliminary}
Unlike prior studies that rely on raw, unconstrained hidden states, we conducted two preliminary experiments addressing the fundamental question: \textit{what is the reasonable representation for latent tokens?}

\subsection{The Distributional Discrepancy in Vector Spaces}
Most existing studies \citep{DBLP:journals/corr/abs-2502-21074,DBLP:journals/corr/abs-2506-18582} follow the COCONUT design, using the last-layer hidden state directly as the representation of the next latent token rather than sampling from the explicit token embeddings. However, a statistical analysis of these vector spaces in representative LLMs reveals a critical discrepancy.

\textbf{Observation 1.} \textit{Significant distributional divergence exists between the last-layer hidden states and the token embedding space.} As illustrated in Figure~\ref{fig_hidden_vs_embeddings} and Figure~\ref{fig_hidden_vs_embeddings_deepseek}, PCA projections demonstrate a substantial separation between the manifolds of hidden states and token embeddings. Quantitative analysis confirms this divergence, showing significant disparities in first and second-order statistics (mean and variance), as well as large inter-distribution distances. From a model optimization perspective, shallow parameters (e.g., RMSNorm) are calibrated exclusively on the token embedding distribution during pre-training. Consequently, introducing vectors from an alien distribution (the raw hidden states) can precipitate distribution shift and training instability, offering a potential explanation for the performance limitations observed in methods such as COCONUT.

\subsection{The Low Effective Rank of Vocab-Space}

Building on Observation 1, we posit that latent token should ideally preserve the distributional geometry of vocab-space.  To investigate the intrinsic dimensionality of this space, we analyze the singular value spectra of token embedding matrices across various LLMs.

\textbf{Observation 2.} \textit{The Token Embedding Matrix exhibits a non-full effective rank.} The singular value decay curves presented in Figure~\ref{fig_svd_decay} exhibit a precipitous drop rather than a gradual plateau, indicating that the vocab-space is dominated by a small subset of singular directions. Despite the high nominal dimensionality of the embedding matrix, its effective rank is considerably lower than the theoretical maximum. This suggests that the semantic manifold of LLMs is inherently low-dimensional.

These findings imply that latent tokens should not be modeled as unconstrained continuous vectors, but rather as structured elements constrained to the embedding subspace. Motivated by this geometric insight, we propose the following formal definition.

\textbf{Definition 1} (\textit{Superposition Embedding})\textbf{.}
We define a latent token $z \in \mathbb{R}^d$ as a vector residing within the linear span of the vocab-space. Formally, let $\mathbf{E} = [e_1, e_2, \dots, e_V] \in \mathbb{R}^{d \times V}$ denote the token embedding matrix, where $e_i \in \mathbb{R}^d$ represents the embedding vector for the $i$-th token in a vocabulary of size $V$. A latent token $z$ is expressed as a linear combination of these embeddings:
\begin{equation}
\label{eq:soft_embedding}
    z = \mathbf{E} \boldsymbol{\alpha} = \sum_{i=1}^V \alpha_i e_i,
\end{equation}
where $\boldsymbol{\alpha} = [\alpha_1, \dots, \alpha_V]^\top \in \mathbb{R}^V$ represents the coefficient vector governing the semantic mixture. This formulation ensures that latent tokens remain aligned with the intrinsic semantic geometry of the LLM’s vocab-space.



\begin{figure*}[t]
  \centering
  \includegraphics[width=\linewidth]{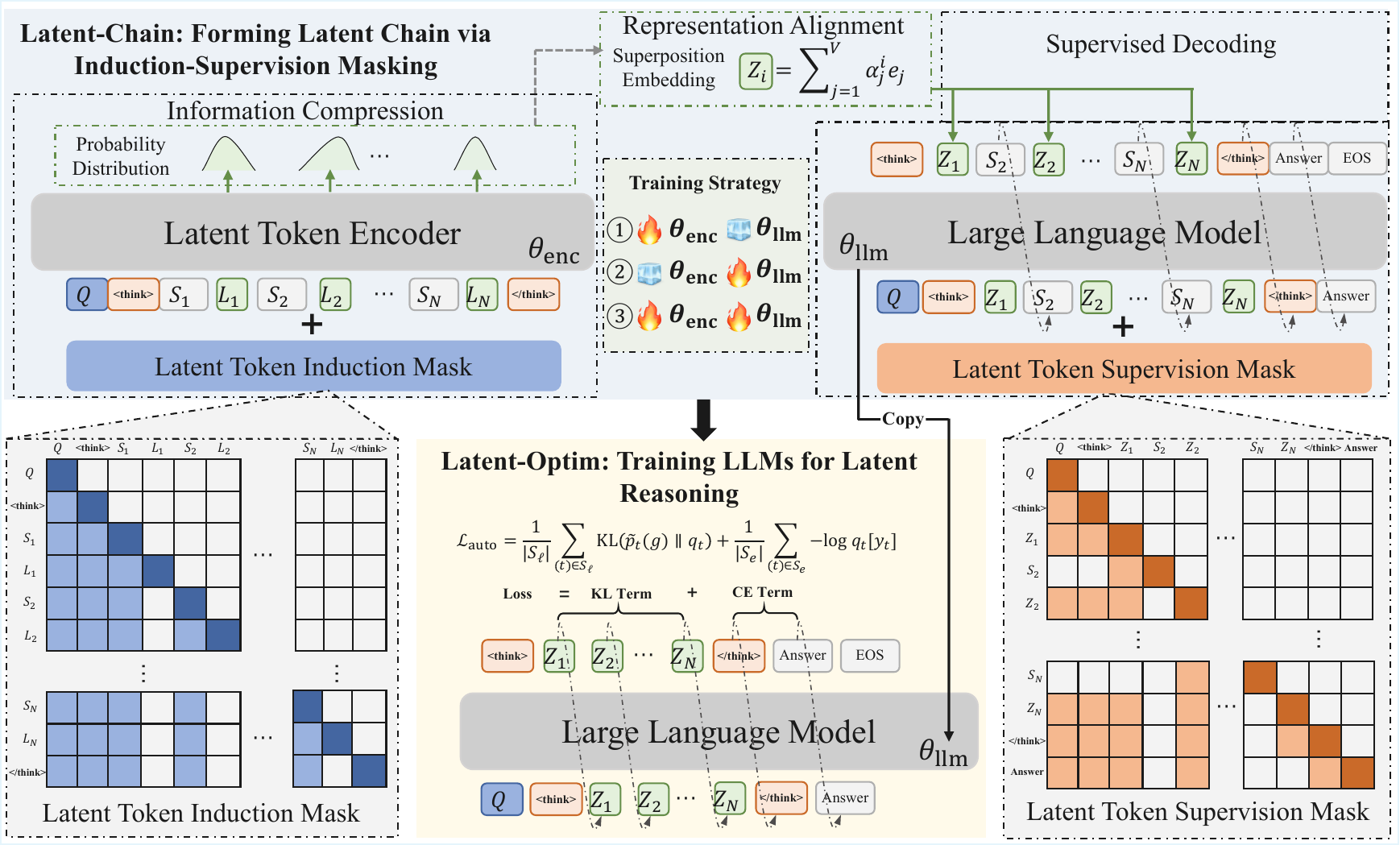}
  \caption{Overview of Latent-SFT. The entire training process consists of two phases: (a) Generating Latent Chains via Induction–Supervision Masking. In the mask sub-figure, components such as $Q$, $S_i$ and Answer (which may span multiple tokens) are implemented via a standard autoregressive attention mechanism. The Latent Token Encoder shares both structure and initialization with the base LLM. (b) Training the LLM to Autonomously Generate Latent Chains. In this phase, the LLM is trained to perform latent reasoning independently, using a weighted combination of KL loss and CE loss.}
  \label{fig_overview}
  \vspace{-0.5cm}
\end{figure*}

\section{Latent-Chain: Forming Latent Chain via Induction-Supervision Masking}
\label{sec:latent_chain}

Having established the definition of the Latent-Vocab, we address the next core question: \textit{what are the requisite properties of the higher-level Latent-Chain?} Unlike prior approaches that treat the latent chains as an ambiguous ``black box'' optimized solely for final answer reconstruction, thereby neglecting intermediate semantic fidelity, we aim to capture the step-by-step logic essential for latent reasoning by postulating that the Latent-Chain must satisfy three key semantic criteria: (1) Semantic Compactness, compressing multiple explicit tokens into a single latent representation; (2) Semantic Compatibility, adhering to the intrinsic semantic space of explicit tokens (as per Section~\ref{sec_preliminary}); and (3) Semantic Sufficiency, retaining sufficient information to deduce the correct answer. As shown in the Figure~\ref{fig_overview}, we implement these properties within a encoder-decoder framework through three key steps: information compression, representation alignment, and supervised decoding.

\paragraph{Information Compression Process (Semantic Compactness).} We insert special tokens into the explicit reasoning sequence to segment and capture intermediate reasoning information. Given an input $X=[Q, \texttt{<think>}, C, \texttt{</think>}]$, we define a segmentation function $\mathrm{Seg}(\cdot)$ that partitions the explicit reasoning chain $C$ into $N$ subsegments, denoted as $C=\{S_i\}_{i=1}^N$. This segmentation can be governed either by a fixed compression ratio $r$ or by semantic boundaries. Following each subsegment $S_i$, we insert a special latent placeholder $L_i$. Crucially, we apply a \textit{Latent Token Induction Mask} (LTIM) to restrict the attention of $L_i$ exclusively to the preceding explicit reasoning context (see Figure~\ref{fig_overview}). This design ensures that the semantic content encoded by each latent token represents a cumulative summary of the reasoning process up to that point.

\paragraph{Representation Alignment Process (Semantic Compatibility).}
For each inserted placeholder $L_i$, we extract its hidden state $\mathbf{h}_i \in \mathbb{R}^d$ from the model under LTIM masking. To strictly enforce \textbf{Semantic Compatibility}, we project $\mathbf{h}_i$ onto the vocabulary simplex and reconstruct it as a mixture of token embeddings. Let $\mathbf{E}=[\mathbf{e}_1, \dots, \mathbf{e}_V] \in \mathbb{R}^{d\times V}$ denote the token embedding matrix and $\mathbf{W} \in \mathbb{R}^{d\times V}$ be the language modeling head (where $\mathbf{W}=\mathbf{E}$ under weight tying). We compute the vocabulary logits and the resulting probability distribution as:
\begin{equation}
\label{eq:rep-align-logits}
\boldsymbol{\ell}_i = \mathbf{W}^{\!\top}\mathbf{h}_i, \quad
\boldsymbol{\alpha}_i = \mathrm{softmax}\! \left(\frac{\boldsymbol{\ell}_i}{\tau}\right) \in \Delta^{V-1},
\end{equation}
where $\tau > 0$ is the temperature parameter. The final latent representation is then obtained via linear interpolation:
\begin{equation}
\label{eq:rep-align-z}
\mathbf{z}_i = \mathbf{E}\boldsymbol{\alpha}_i = \sum{v=1}^V \alpha_{i,v}\mathbf{e}_v.
\end{equation}
This operation instantiates Definition 1, guaranteeing that $\mathbf{z}_i$ resides within the convex hull of the embedding space $\{\mathbf{e}_v\}_{v=1}^V$. Consequently, latent tokens remain confined to the intrinsic geometry of the LLM’s vocabulary manifold. During the subsequent optimization phase, the distribution $\boldsymbol{\alpha}_i$ serves as the supervision target (via KL divergence) for the latent positions.

\paragraph{Supervised Decoding Process (Semantic Sufficiency).}
Given the latent representations $\{\mathbf{z}_i\}_{i=1}^N$ from the alignment step, we use each $\mathbf{z}_i$ to decode the content from the ($i\!+\!1$)-th explicit reasoning subsegment through to the final answer. Let the prefix up to step $i$ be $\Pi_i = [Q, \texttt{<think>}, \mathbf{z}_1, S_2, \mathbf{z}_2, \dots, \mathbf{z}_i]$. Under the Latent Token Supervision Mask (LTSuM), tokens in $\mathcal{Y}_{i} = [S_{i+1}, \dots, S_N, \texttt{</think>}, \mathrm{Answer}]$ are decoded conditioned only on $\Pi_i$ (in particular on $[Q,\texttt{<think>},\mathbf{z}_1,...,\mathbf{z}_i]$), while attention to future latent tokens and to past explicit reasoning steps is blocked. Denoting by $\mathcal{J}_i$ the index set of tokens in $\mathcal{Y}_{i}$, the supervised decoding objective is
\begin{equation}
\label{eq:sup-dec}
\mathcal{L}_{\mathrm{sup}} \;=\; \frac{1}{N}\sum_{i=1}^{N} \frac{1}{|\mathcal{J}_i|} \sum_{t\in\mathcal{J}_i} \Big(-\log p_\theta\big(x_t \,\big|\, \Pi_i;\, \text{LTSuM}\big)\Big).
\end{equation}

\paragraph{Training Procedure.} 

The strict constraints imposed by LTIM and LTSuM can destabilize naive SFT optimization. To address this, we adopt an EM-style alternating optimization schedule. Let $\theta = \{\theta_{\mathrm{enc}}, \theta_{\mathrm{llm}}\}$ represent the parameters, where $\theta_{\mathrm{enc}}$ denotes the parameters active during latent token induction. The training proceeds in three phases:
\begin{enumerate}
\item \textbf{Encoder Optimization:} Freezing the LLM backbone, we update $\theta_{\mathrm{enc}}$ to minimize reconstruction loss: $\theta_{\mathrm{enc}}^{(t+1)} \leftarrow \mathrm{arg}\min_{\theta_{\mathrm{enc}}} \mathcal{L}_{\mathrm{sup}}(\theta_{\mathrm{enc}}, \theta_{\mathrm{llm}}^{(t)})$.
\item \textbf{LLM Optimization:} Freezing the encoder, we update the backbone to adapt to the latent representations: $\theta_{\mathrm{llm}}^{(t+1)} \leftarrow \mathrm{arg}\min_{\theta_{\mathrm{llm}}} \mathcal{L}_{\mathrm{sup}}(\theta_{\mathrm{enc}}^{(t+1)}, \theta_{\mathrm{llm}})$.
\item \textbf{Joint Fine-tuning:} Finally, we jointly optimize both components to ensure global convergence.
\end{enumerate}
This progressive strategy stabilizes the training dynamics, allowing the model to gradually adapt to the distributional shift introduced by the latent tokens.

\section{Latent-Optim: Training LLMs for Latent Reasoning}
\label{sec:latent_sft}

Having constructed the Latent-Chain with the requisite semantic properties in Section \ref{sec:latent_chain}, we address the final question: \textit{how can we empower the LLM to generate these high-quality latent chains during reasoning autonomously?} Unlike existing methods that rely on inefficient indirect backpropagation from final supervision—thereby lacking explicit guidance for the intermediate latent process—we propose Latent-Optim, a training strategy to learn robust reasoning chains while ensuring generalization.

We concatenate latent chains with the explicit context to form the training sequence $X = [Q, \texttt{<think>}, \mathbf{z}_1, \dots, \mathbf{z}_N, \texttt{</think>}, \mathrm{Answer}]$. A straightforward optimization approach is to minimize the KL divergence between the static latent chain $\boldsymbol{\alpha}_t$ derived from the encoder and the model's predicted distribution $\mathbf{q}_t = \mathrm{softmax}(\mathbf{W}^\top \mathbf{h}_t)$, combined with the standard cross-entropy loss for explicit tokens:
\begin{equation}
\label{eq:naive_loss}
\begin{split}
\mathcal{L}_{\mathrm{static}} &= \frac{1}{|S_{\mathrm{lat}}|}\sum_{t\in S_{\mathrm{lat}}} \mathrm{KL}(\boldsymbol{\alpha}_t \,||\, \mathbf{q}_t) \\
&\quad + \frac{1}{|S_{\mathrm{exp}}|}\sum_{t\in S_{\mathrm{exp}}} \big(-\log q_t[y_t]\big).
\end{split}
\end{equation}

However, directly supervising the LLM with static latent chains $\boldsymbol{\alpha}_t$ can lead to overfitting on the specific artifacts of the encoder, thereby limiting the model's generalization capability.

To address this, we propose a \textbf{Stochastic Latent Optimization} strategy. Instead of treating the latent targets as immutable labels, we view them as parameters of a generative distribution. During each training epoch, we dynamically sample a perturbed target distribution using the Gumbel-Softmax trick \citep{wu2025llmssinglethreadedreasonersdemystifying}. Specifically, let $S_{\mathrm{lat}}$ and $S_{\mathrm{exp}}$ denote the index sets for latent and explicit positions, respectively. For a latent position $t \in S_{\mathrm{lat}}$, let $\boldsymbol{\alpha}_t$ be the static latent chain. We construct a stochastic target $\tilde{\mathbf{p}}_t$ by injecting Gumbel noise:
\begin{equation}
\label{eq:gumbel_target}
    \tilde{\mathbf{p}}_t = \mathrm{softmax}\left( \frac{\log \boldsymbol{\alpha}_t + \mathbf{g}_t}{\tau} \right), \, \mathbf{g}_t \sim \mathrm{Gumbel}(0, 1),
\end{equation}
in our experiments, we set the temperature $\tau=1$, which simplifies the unperturbed expectation to the original target $\boldsymbol{\alpha}_t$ while introducing crucial stochasticity. The noise vector $\mathbf{g}_t$ is resampled independently at every training step, ensuring that the model is never exposed to the exact same latent target twice. This dynamic perturbation forces the model to learn the underlying semantic manifold rather than memorizing specific point estimates. 

Consequently, the final optimization objective is reformulated as the expectation over this noise:
\begin{equation}
\label{eq:llm-auto}
\begin{split}
\mathcal{L}_{\mathrm{auto}}(\theta_{\mathrm{llm}})
&= \frac{1}{|S_{\mathrm{lat}}|}\sum_{t\in S_{\mathrm{lat}}} \mathbb{E}_{\mathbf{g}}\left[ \mathrm{KL}\big(\tilde{\mathbf{p}}_t(\mathbf{g}) \,||\, \mathbf{q}_t\big) \right] \\
&\quad + \frac{1}{|S_{\mathrm{exp}}|}\sum_{t\in S_{\mathrm{exp}}} \big(-\log q_t[y_t]\big).
\end{split}
\end{equation}
By minimizing the expectation over the Gumbel noise, we effectively smooth the optimization landscape. As we formally prove in \textbf{Theorem 1} (see Appendix \ref{app:proof}), this stochastic injection acts as an implicit regularizer that penalizes the sharpness of the loss Hessian, thereby guiding the model toward flatter minima and enhancing generalization across diverse reasoning paths.

\begin{table*}[t]
\centering
\tiny
\caption{Experimental results on four low-difficulty datasets. The number in brackets represents the compression ratio $r$. “\textcolor{bestbg}{\rule{5mm}{2mm}}” marks the best result, and “\textcolor{secondbg}{\rule{5mm}{2mm}}” marks the second best. \colorbox[HTML]{E0E0E0}{~~-~Hidden State} refers to removing the Representation Alignment step and directly using the last-layer hidden state as the representation of latent tokens. \colorbox[HTML]{E0E0E0}{~~-~w/o Gumbel} indicates removing the stochastic Gumbel noise injection, effectively performing deterministic optimization on static soft labels. \colorbox[HTML]{E0E0E0}{~~-~w/o LTIM} and \colorbox[HTML]{E0E0E0}{~~-~w/o LTSuM} denote removing the Latent Token Induction Mask and Latent Token Supervision Mask, respectively, reducing the model to a standard autoregressive attention mechanism.}
\label{tab_low_difficulty}
\resizebox{\linewidth}{!}{%
\begin{tabular}{l|cc|cc|cc|cc|ccc}
\toprule
        & \multicolumn{2}{c|}{GSM8k-Aug} & \multicolumn{2}{c|}{GSM-Hard} & \multicolumn{2}{c|}{SVAMP} & \multicolumn{2}{c|}{MultiArith} & \multicolumn{3}{c}{Average} \\
        & Pass@1 $\uparrow$  & \#~L $\downarrow$  & Pass@1 $\uparrow$         & \#~L $\downarrow$          & Pass@1 $\uparrow$        & \#~L $\downarrow$        & Pass@1 $\uparrow$          & \#~L $\downarrow$          & Pass@1 $\uparrow$           & \#~L $\downarrow$ &Pass@1/\#~L $\uparrow$\\ \midrule
\multicolumn{12}{c}{\textit{LLaMA-3.2-1B-Instruct}} \\
\midrule
CoT-SFT     & \second{49.4$_{\pm.6}$}    & 70.1$_{\pm.2}$   & \second{12.1$_{\pm.2}$}    & 78.4$_{\pm.1}$    & \second{60.5$_{\pm.4}$}   & 37.2$_{\pm.1}$  & 93.2$_{\pm.5}$     & 38.5$_{\pm.1}$   & \second{53.8}     & 56.5 &0.95\\ \midrule
iCoT   & 19.8$_{\pm.2}$    & 0.00$_{\pm.0}$         & 3.87$_{\pm.2}$    & 0.00$_{\pm.0}$          & 36.4$_{\pm.5}$   & 0.00$_{\pm.0}$        & 38.2$_{\pm.7}$     & 0.00$_{\pm.0}$          & 24.6     & 0.00 & - \\
COCONUT & 23.1$_{\pm.3}$    & 6.00$_{\pm.0}$          & 5.49$_{\pm.3}$    & 6.00$_{\pm.0}$           & 40.7$_{\pm.7}$   & 6.00$_{\pm.0}$         & 41.1$_{\pm.2}$     & 6.00$_{\pm.0}$          & 27.6    & 6.00 & 4.60\\
CODI  & 13.3$_{\pm.6}$    & 6.00$_{\pm.0}$          & 2.97$_{\pm.2}$    & 6.00$_{\pm.0}$           & 21.7$_{\pm.7}$   & 6.00$_{\pm.0}$         & 19.2$_{\pm.8}$     & 6.00$_{\pm.0}$          & 14.3     & 6.00& 2.38\\ 
CoLaR-5 & 26.8$_{\pm.2}$    & 5.57$_{\pm.0}$   & 5.87$_{\pm.1}$    & 6.53$_{\pm.0}$    & 48.4$_{\pm.5}$   & 2.95$_{\pm.0}$  & 86.4$_{\pm.4}$     & 3.21$_{\pm.0}$    & 41.7     & 4.57 &\second{9.12}\\
CoLaR-2 & 40.1$_{\pm.2}$    & 12.7$_{\pm.0}$   & 9.08$_{\pm.0}$    & 14.0$_{\pm.0}$    & 54.9$_{\pm.2}$   & 6.11$_{\pm.0}$  & 91.3$_{\pm.1}$     & 7.35$_{\pm.0}$    & 48.8     & 10.0 & 4.88\\ \midrule
Latent-SFT(4) & 46.8$_{\pm.1}$    & 6.81$_{\pm.0}$   & 11.2$_{\pm.2}$    & 7.77$_{\pm.2}$    & 57.6$_{\pm.3}$   & 3.60$_{\pm.1}$  &  \second{94.0$_{\pm.2}$}     & 4.15$_{\pm.0}$    & 52.4     & 5.58 &\best{9.39} \\
\rowcolor[HTML]{E0E0E0}
~~-~w/o Gumbel & 44.6$_{\pm.0}$    & 6.64$_{\pm.0}$   & 10.2$_{\pm.0}$    & 7.68$_{\pm.0}$    & 55.6$_{\pm.1}$   & 3.33$_{\pm.0}$  &  93.6$_{\pm.0}$     & 4.11$_{\pm.0}$    & 51.0     & 5.44 &9.37 \\
\rowcolor[HTML]{E0E0E0}
~~-~Hidden State & 36.7$_{\pm.4}$    & 6.61$_{\pm.1}$   & 8.41$_{\pm.2}$    & 7.77$_{\pm.0}$    & 44.4$_{\pm.2}$   & 3.22$_{\pm.0}$  & 90.3$_{\pm.1}$     & 4.06$_{\pm.0}$    & 45.0     & 5.41 & 8.32 \\
\rowcolor[HTML]{E0E0E0}
~~-~w/o LTIM & 38.8$_{\pm.1}$    & 6.59$_{\pm.1}$   & 8.67$_{\pm.1}$    & 8.01$_{\pm.1}$    & 48.9$_{\pm.1}$   & 3.34$_{\pm.1}$  & 90.8$_{\pm.0}$     & 4.36$_{\pm.0}$    & 46.8     & 5.58 & 8.39 \\
\rowcolor[HTML]{E0E0E0}
~~-~w/o LTSuM & 42.6$_{\pm.0}$    & 6.70$_{\pm.0}$   & 9.98$_{\pm.1}$    & 7.60$_{\pm.0}$    & 53.9$_{\pm.1}$   & 3.36$_{\pm.0}$  & 92.6$_{\pm.0}$     & 4.27$_{\pm.0}$    & 49.7     & 5.48 & 9.07 \\
\midrule
Latent-SFT(2) & \best{52.4$_{\pm.2}$}   & 12.8$_{\pm.0}$   & \best{12.6$_{\pm.4}$}    & 15.0$_{\pm.1}$    & \best{61.3$_{\pm.4}$}   & 6.21$_{\pm.1}$  & \best{96.8$_{\pm.3}$}     & 7.19$_{\pm.2}$    & \best{55.8}\accgain{2.0}     & 10.3\speedup{5.5} &5.41\\
\rowcolor[HTML]{E0E0E0}
~~-~w/o Gumbel & 50.4$_{\pm.1}$   & 12.4$_{\pm.0}$   & 10.9$_{\pm.0}$    & 13.9$_{\pm.0}$    & 57.8$_{\pm.1}$   & 5.98$_{\pm.0}$  & 93.8$_{\pm.0}$     & 7.21$_{\pm.0}$    & 53.2     & 9.87 &5.39\\
\rowcolor[HTML]{E0E0E0}
~~-~Hidden State & 41.1$_{\pm.3}$    & 12.8$_{\pm.0}$   & 9.01$_{\pm.1}$    & 14.2$_{\pm.1}$    & 49.6$_{\pm.2}$   & 6.17$_{\pm.0}$  & 91.1$_{\pm.0}$     & 7.29$_{\pm.0}$    & 47.7     & 10.1 & 4.72 \\
\rowcolor[HTML]{E0E0E0}
~~-~w/o LTIM & 41.3$_{\pm.1}$    & 12.9$_{\pm.1}$   & 9.32$_{\pm.0}$    & 14.0$_{\pm.0}$    & 53.6$_{\pm.1}$   & 6.25$_{\pm.1}$  & 91.5$_{\pm.0}$     & 7.96$_{\pm.0}$    & 48.9     & 10.3 & 4.75 \\
\rowcolor[HTML]{E0E0E0}
~~-~w/o LTSuM & 49.0$_{\pm.0}$    & 12.1$_{\pm.0}$   & 10.3$_{\pm.0}$    & 13.7$_{\pm.0}$    & 56.5$_{\pm.2}$   & 6.22$_{\pm.0}$  & 93.0$_{\pm.0}$     & 7.23$_{\pm.0}$    & 52.2     & 9.81 & 5.32 \\
\midrule
\multicolumn{12}{c}{\textit{LLaMA-3.2-3B-Instruct}} \\
\midrule
CoT-SFT     & \second{61.1$_{\pm.3}$}    & 70.5$_{\pm.2}$   & \best{18.8$_{\pm.2}$}    & 80.6$_{\pm.1}$    & \second{70.9$_{\pm.2}$}   & 34.2$_{\pm.1}$  & \best{98.3}$_{\pm.5}$     & 37.4$_{\pm.1}$   & \second{62.2}     & 55.7 &\second{1.12}\\ 
Latent-SFT(2) & \best{62.4$_{\pm.1}$}   & 12.5$_{\pm.3}$   & \second{18.3$_{\pm.3}$}    & 14.6$_{\pm.2}$    & \best{71.3$_{\pm.7}$}   & 5.94$_{\pm.1}$  & \second{98.7$_{\pm.4}$}     & 7.15$_{\pm.1}$    & \best{62.7}\accgain{0.5}     & \hspace{1.2em}10.3\speedup{5.6}\hspace{1.2em} &\best{5.39}\\

\bottomrule
\end{tabular}
}
\vspace{-0.5cm}
\end{table*}
\section{Experiments}
\subsection{Experimental Setups}
\paragraph{Datasets.} To comprehensively assess the effectiveness of our method, we conducted experiments on both low-difficulty mathematical reasoning datasets (GSM8k-Aug \citep{DBLP:journals/corr/abs-2405-14838}, GSM-Hard \citep{DBLP:conf/icml/GaoMZ00YCN23}, SVAMP \citep{DBLP:conf/naacl/PatelBG21}, and MultiArith \citep{DBLP:conf/emnlp/RoyR15}) and high-difficulty datasets (Math500 \citep{DBLP:conf/nips/HendrycksBKABTS21} and AIME24). For the low-difficulty tasks, we trained on the GSM8k-Aug training split to ensure fair comparison with prior work. For the high-difficulty tasks, due to limited computational resources, we used a subset of Open-R1 \citep{openr1} containing explicit reasoning chains shorter than 4k tokens, resulting in 57,544 training samples. This restriction inevitably constrained the performance of our method. 

\paragraph{Baselines and Models.} The baselines considered in our experiments include COT-SFT \citep{DBLP:conf/nips/Wei0SBIXCLZ22}, iCOT \citep{DBLP:journals/corr/abs-2405-14838}, COCONUT \citep{DBLP:journals/corr/abs-2412-06769}, CODI (reproduced) \citep{DBLP:journals/corr/abs-2505-16552}, and CoLaR \citep{DBLP:journals/corr/abs-2505-16552}. To align model capabilities with task complexity, we strictly match base models to dataset difficulty. For low-difficulty tasks, we adopt LLaMA-3.2-1B-Instruct and LLaMA-3.2-3B-Instruct \citep{DBLP:journals/corr/abs-2407-21783}, while for high-difficulty tasks, we employ DeepSeek-Distill-Qwen-7B \citep{DBLP:journals/corr/abs-2501-12948, yang2025qwen3technicalreport}. 

\paragraph{Evaluation Metrics and Details.} Our evaluation measures both accuracy (Pass@N) and efficiency (\#~L, the number of tokens in the reasoning chain). For the low-difficulty datasets, we conducted five runs with different random seeds. For the high-difficulty datasets, Math500 was evaluated four times, while AIME24 was evaluated 64 times. We report Pass@1 and \#~L as the mean ± 95\% confidence interval (CI). To jointly assess accuracy and efficiency, we also report their ratio, Pass@1/\#~L. 

\paragraph{Implementation Details} Since Latent-SFT requires weighting with respect to the vocabulary matrix, we freeze the vocabulary parameters during training. All models are trained using LoRA with rank 64. Following COCONUT, we initialize all methods from the COT-SFT model to accelerate convergence. In practice, we find that a fixed-length segmentation function performs better than a semantic-based segmentation. Therefore, all results reported in this paper are based on fixed-length segmentation functions. For LLaMA-3.2-1B-Instruct and LLaMA-3.2-3B-Instruct, we set the learning rate to $1\times 10^{-4}$for Latent-Chain generation and $4\times 10^{-4}$ for Latent-Optim. For DeepSeek-Distill-Qwen-7B, the learning rates are $3\times 10^{-5}$ (Latent-Chain) and $1\times 10^{-5}$ (Latent-Optim). Following \cite{DBLP:journals/corr/abs-2501-12948}, we use a decoding temperature of 0.6 and top-p of 0.95. 

During evaluation, low-difficulty mathematical reasoning datasets were inferred using the Transformers library with a maximum sequence length of 128. High-difficulty datasets were evaluated with the improved SGlang library, supporting a maximum token length of 32,768. All training and inference experiments were conducted on 8×A100 GPUs.

\begin{table*}[t]
\centering
\tiny
\caption{Experimental results of DeepSeek-Distill-Qwen-7B on high-difficulty datasets. We primarily examine the performance gap between \textit{Superposition Embedding} and hidden state representations. To ensure a fair comparison, we set the maximum reasoning chain length to 4K tokens for all models.}
\label{tab_hign_difficulty}
\resizebox{\linewidth}{!}{
\begin{tabular}{l|ccc|ccc}
  \toprule
  \multirow{2}{*}{Model} & \multicolumn{3}{c}{MATH-500} & \multicolumn{3}{c}{AIME24} \\
  \cmidrule(lr){2-4} \cmidrule(lr){5-7}
   & Pass@1 $\uparrow$ & \#~L $\downarrow$ & Pass@1/\#~L$\times 100$ $\uparrow$ & Pass@1 $\uparrow$ & \#~L $\downarrow$ & Pass@1/\#~L$\times 100$ $\uparrow$\\ \hline
  COT-SFT & 73.8$_{\pm1.32}$ & 2766$_{\pm319}$ & 2.67 & 17.7$_{\pm1.23}$ & 4016$_{\pm343}$ & 0.44 \\ \hline
  Latent-SFT (Hidden State) & 67.8$_{\pm0.94}$ & 698.1$_{\pm22.1}$ & 9.71 & 7.78$_{\pm2.08}$ & 1509$_{\pm114}$ & 0.51 \\
  Latent-SFT (\textit{Superposition Embedding}) & \textbf{79.8}$_{\pm1.35}$\accgain{6.0} & 759.3$_{\pm36.9}$\speedup{3.6} & \textbf{10.51} & \textbf{19.2}$_{\pm1.60}$\accgain{1.5} & 1484$_{\pm420}$\speedup{2.7} & \textbf{1.29} \\
  \toprule
\end{tabular}
}
\vspace{-0.5cm}
\end{table*}

\subsection{Main Results}

\paragraph{Performance on Low-Difficulty Tasks.}
Table~\ref{tab_low_difficulty} compares Latent-SFT with state-of-the-art baselines under different compressed ratio $r$ settings. Relative to the latent reasoning baseline, Latent-SFT consistently achieves higher accuracy, shorter reasoning chains, and substantial gains in the Pass@1 metric. These results confirm that defining latent tokens as a superposition of vocab-space—thus aligning them with the same semantic space as explicit tokens—is more effective than the traditional last-hidden-state definition. Furthermore, Latent-SFT demonstrates strong out-of-distribution generalization, achieving state-of-the-art performance on all three out-of-distribution datasets.

Notably, across models of varying sizes, Latent-SFT(2) consistently outperforms explicit CoT-SFT in Pass@1 scores on nearly all mathematical reasoning datasets. This advantage is particularly significant in the 1B model, where it improves average Pass@1 performance by 2.0\% while reducing the average reasoning chain length by approximately 5.5$\times$. This finding demonstrates that explicit reasoning does not represent the upper bound of latent reasoning methods. With an appropriate definition and learning strategy for latent tokens, latent reasoning can exceed explicit reasoning models in accuracy. The real reasoning example is provided in the Figure~\ref{fig_gsm8k_case}.

\paragraph{Performance on High-Difficulty Reasoning Tasks.}
Experimental results on high-difficulty mathematical tasks are shown in Table~\ref{tab_hign_difficulty}. Our method maintains its effectiveness on the 7B model, surpassing explicit CoT-SFT in Pass@1 performance while reducing the reasoning chain length by 2.7--3.6$\times$. This result is particularly significant given that prior latent reasoning methods often suffer from semantic collapse when handling intricate, multi-step deductions (e.g., AIME). In contrast, Latent-SFT demonstrates remarkable robustness, effectively compressing complex logical trajectories without losing critical intermediate information. More importantly, \textit{Superposition Embedding}, the core definition proposed in this work, consistently outperforms Hidden State even on high-difficulty datasets, thereby validating the generalizability of \textbf{Definition 1}. This confirms that constraining latent reasoning to the vocabulary manifold is essential for scaling to competition-level mathematics, where unconstrained continuous vectors struggle to maintain precise semantic alignment.

\subsection{Ablation Study}

\paragraph{Necessity of Masking Mechanisms.} 
LTIM enforces causal independence to stabilize training, while LTSuM ensures latent tokens retain sufficient semantic information for decoding. As shown in Table~\ref{tab_low_difficulty}, removing either mechanism leads to a significant accuracy drop ($\sim$5\%), confirming their critical role in maintaining the structural integrity of the reasoning chain.

\paragraph{Latent Representation and Stochastic Optimization.} 
We further validate our specific design choices. First, replacing \textit{Superposition Embedding} with raw hidden states degrades performance, demonstrating the importance of aligning latent tokens with the intrinsic vocabulary manifold. Second, ablating Gumbel-noise injection (w/o Gumbel) leads to a noticeable decline, confirming that stochasticity acts as a vital regularizer (Theorem 1) to prevent overfitting and enhance generalization.
\begin{figure}[t]
  \centering
  \includegraphics[width=\linewidth]{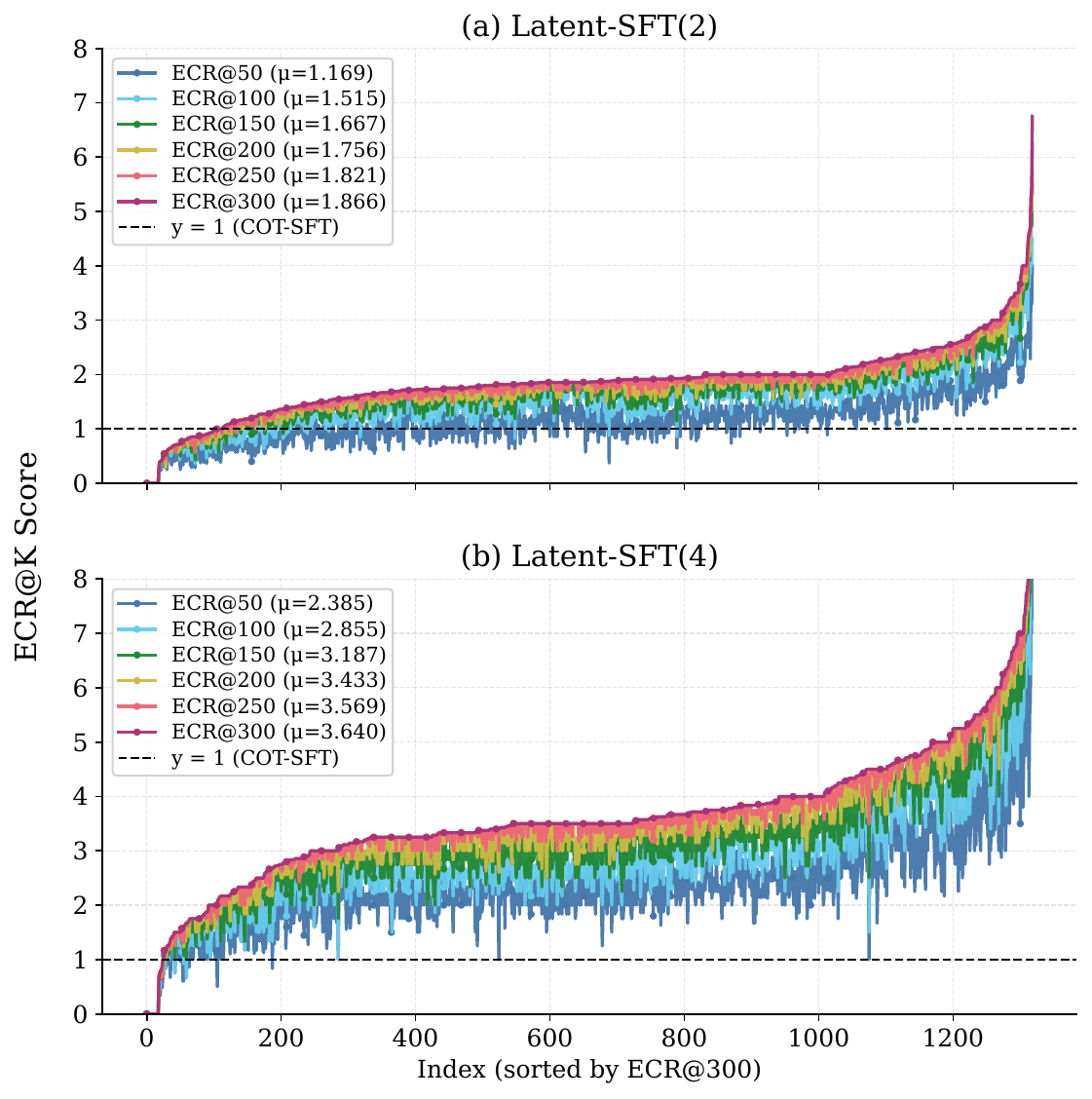}
  \caption{ECR@K analysis of Latent-SFT on the GSM8k-Aug test set. Each curve represents the per-sample ECR@K, with samples sorted along the x-axis by their ECR@300 values. The dashed line indicates the baseline for uncompressed single-chain reasoning. Latent-SFT consistently surpasses this baseline, with higher training compression ratios yielding greater information compression. This demonstrates that Latent-SFT effectively compresses single explicit reasoning paths during the latent reasoning.}
  \label{fig_llama2_ecr}
  \vspace{-0.5cm}
\end{figure}

\begin{figure}[t]
  \centering
  \includegraphics[width=0.8\linewidth]{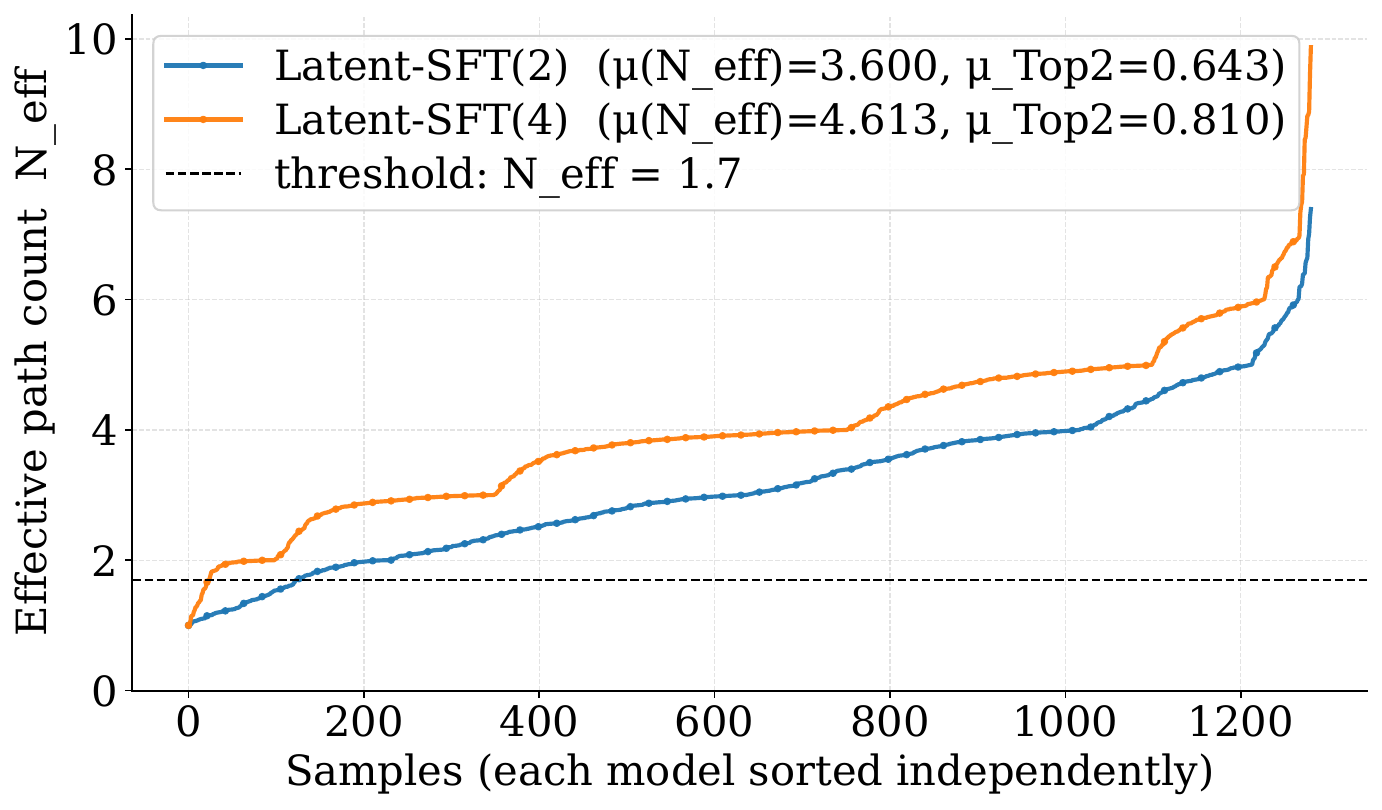}
  \caption{$N_{\mathrm{eff}}$ values of Latent-SFT on the Multi-Chain GSM8k-Aug dataset. The legend reports the mean values of both $N_{\mathrm{eff}}$ and the Top-2 Score. Overall, Latent-SFT models across various compression ratios consistently exceed the significant parallel threshold. This confirms that the latent reasoning process effectively operates as a superposition of multiple simultaneous reasoning chains.}
  \label{fig_neff}
  \vspace{-0.3cm}
\end{figure}

\begin{figure}[t]
  \centering
  \includegraphics[width=\linewidth]{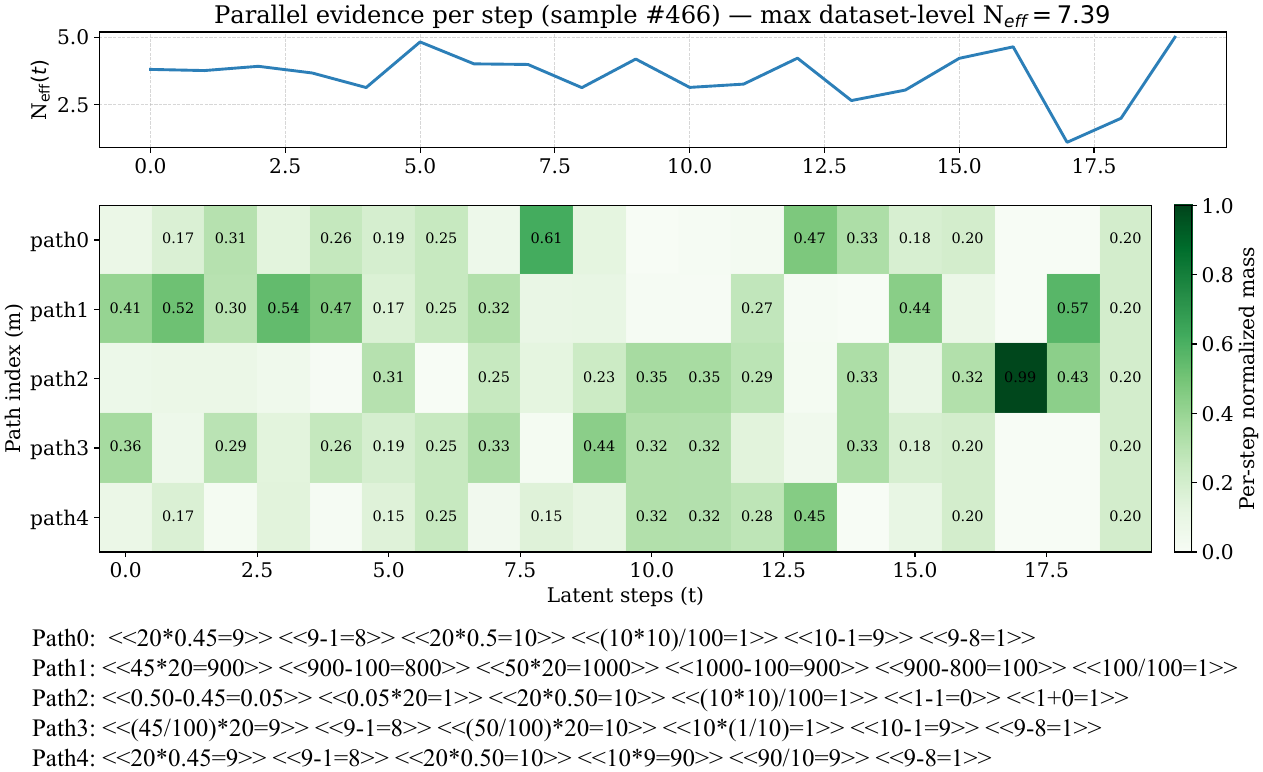}
  \caption{An example of the maximum N$_{\mathrm{eff}}$ value for the Latent-SFT(2) model on the Multi-Chain GSM8k-Aug dataset. At each step, the  N$_{\mathrm{eff}}$ value exceeds 1.7. Path0 denotes the original reasoning chain, while the remaining paths are additional candidate chains. The heatmap illustrates the fraction of paths supported at each latent step, clearly revealing the phenomenon of parallel reasoning.}
  \label{fig_infer_case}
  \vspace{-0.3cm}
\end{figure}
\section{Essence of Latent Reasoning}
A central premise of our proposed Latent-SFT is that it enables the model to transcend simple sequence compression and instead operate in a state of \textit{reasoning superposition}. This raises a fundamental validation question: \textit{Has our method truly realized this ideal state of latent reasoning?} Specifically, does the learned latent process effectively capture and explore multiple reasoning paths simultaneously, rather than merely memorizing a single trajectory? To rigorously answer this, we investigate the underlying mechanism of Latent-SFT through two hypotheses.

\paragraph{Hypothesis 1. Latent Reasoning as the Compression of a Single Reasoning Chain.}
To quantify how much explicit information is preserved at each step of latent reasoning, we introduce the \textbf{Top-K Effective Compression Ratio (ECR@K)}. This metric measures the proportion of explicit chain tokens that fall within the Top-K of the probability distribution at the corresponding latent step. Larger values indicate stronger horizontal compression across the reasoning chain, while values greater than one reflect true compression of the explicit chain. The formal definition is provided in the Appendix~\ref{app_ecr}.

The experimental results are presented in Figure~\ref{fig_llama2_ecr}. Across models trained with different compression ratios r and varying K values, the ECR scores exceed 1 for almost all samples. This quantitatively validates our hypothesis that latent reasoning compresses a single reasoning path. From the training perspective, the model indeed performs information compression along a single reasoning path. Moreover, comparing Figure~\ref{fig_llama2_ecr}(a) and (b), doubling the training compression ratio nearly doubles the ECR score during inference, further demonstrating the effectiveness of our training strategy.

\paragraph{Hypothesis 2. Latent Reasoning as the Superposition of Multiple Reasoning Chains.}
To examine whether latent reasoning corresponds to the superposition of multiple paths, we define the effective global parallelism (N$_{\mathrm{eff}}$), which quantifies global evidence for the presence of multiple parallel reasoning paths. A larger value reflects a higher degree of parallelism. In addition, we introduce the Top-2 Score, defined as the ratio of the probabilities of the two most likely paths. Formal definitions are provided in the Appendix~\ref{app_neff}. To assess the parallel reasoning capability of latent models, we construct multiple alternative reasoning chains corresponding to a single explicit reasoning path within GSM8k-Aug dataset (Multi-Chain GSM8k-Aug dataset in Appendix~\ref{app_dataset}).

Figure~\ref{fig_neff} presents the analysis results. Across models with different compression ratios, most samples exceed the conservative threshold for significant parallelism (1.7), indicating that latent reasoning reflects the superposition of multiple reasoning chains. On average, evidence is distributed across roughly 3–4 reasoning paths rather than concentrated on a single path. The average Top-2 ratio exceeds 0.6, suggesting strong “two-way competition” and parallel reasoning. Furthermore, as the compression ratio increases, the degree of parallelism also rises. Case studies are provided in the Figure~\ref{fig_infer_case}.

\section{Conclusion}
In this work, we introduce Latent-SFT, a unified framework designed to enable latent reasoning to realize the ideal state of multi-path superposition. By systematically addressing challenges at the token, chain, and learning levels, we transform latent reasoning from an unconstrained "black box" into a structured, interpretable process. Empirical evaluations across six mathematical benchmarks demonstrate that Latent-SFT consistently outperforms explicit SFT while achieving a 2.7x to 5.5x reduction in reasoning chain length. Crucially, our analysis confirms that the learned latent process is not merely a compression of a single explicit path, but a superposition of multiple reasoning chains. This work suggests that the future of efficient LLM reasoning lies not in discarding the chain-of-thought, but in evolving it into a multiple paths, high-entropy latent reasoning.




\section*{Impact Statement}

This paper presents work whose goal is to advance the field of Machine
Learning. There are many potential societal consequences of our work, none
which we feel must be specifically highlighted here.


\bibliography{example_paper}
\bibliographystyle{icml2026}

\newpage
\appendix
\onecolumn

\section{Use of LLMs}
We employ GPT-5 to polish the language of our manuscripts. Specifically, we first draft the text, then use GPT-5 to refine it into a more coherent form. After verifying its logical and semantic consistency, we incorporate the polished version into the main text.

\section{Training Computational Costs}
\label{app:training_cost}

We provide a detailed breakdown of the computational time required for both the \textbf{Latent-Chain Construction} and \textbf{Latent-Optim} phases. All models were trained on 8 $\times$ NVIDIA A100 GPUs.

The training process for Latent-SFT is divided into two primary steps. The first step, \textbf{Latent-Chain Construction}, involves three sub-phases: (1) Encoder Training, (2) Decoder Training, and (3) Joint Fine-tuning. The second step, \textbf{Latent-Optim}, involves training the LLM using the proposed stochastic Gumbel-Softmax strategy.

Table~\ref{tab:training_cost} summarizes the training duration for each model size. Despite the multi-stage nature of our framework, the total training cost remains manageable. 

\begin{table}[h]
    \centering
    \caption{Training time breakdown (in hours) for Latent-SFT across different model scales.}
    \label{tab:training_cost}
    \resizebox{\linewidth}{!}{
    \begin{tabular}{l|cccc|c|c}
        \toprule
        \multirow{2}{*}{\textbf{Base Model}} & \multicolumn{4}{c|}{\textbf{Latent-Chain Construction (h)}} & \multirow{2}{*}{\textbf{Latent-Optim (h)}} & \multirow{2}{*}{\textbf{Total (h)}} \\
        \cmidrule(lr){2-5}
         & \textit{Encoder} & \textit{Decoder} & \textit{Joint} & \textbf{Subtotal} &  &  \\
        \midrule
        LLaMA-3.2-1B-Instruct & 4.0 & 4.0 & 4.5 & 12.5 & 12.0 & 24.5 \\
        LLaMA-3.2-3B-Instruct & 6.5 & 6.5 & 8.0 & 21.0 & 22.0 & 43.0 \\
        DeepSeek-Distill-Qwen-7B & 12.0 & 12.0 & 15.0 & 39.0 & 50.0 & 89.0 \\
        \bottomrule
    \end{tabular}
    }
\end{table}

\section{Theoretical Analysis of Stochastic Latent Optimization}
\label{app:proof}

In this section, we analyze the implicit regularization effect of Gumbel-noise injection. Unlike standard Gaussian noise, Gumbel noise has a non-zero mean and fixed variance. We show that, combined with the translation invariance of Softmax, this induces a Hessian-based regularization term in expectation.

\textbf{Theorem 1} (\textit{Implicit Hessian Regularization via Gumbel-Softmax}). 
\textit{
Let $\boldsymbol{\alpha}\in\Delta^{V-1}$ be a static target distribution in the simplex interior (i.e., $\alpha_i>0$ and $\sum_i\alpha_i=1$).
Consider the stochastic target generated via Eq.~(\ref{eq:gumbel_target}):
$\tilde{\mathbf{p}}=\sigma\!\left((\log\boldsymbol{\alpha}+\mathbf{g})/\tau\right)$,
where $\sigma(\cdot)$ is Softmax, $\tau>0$, and $\mathbf{g}\in\mathbb{R}^V$ has i.i.d. entries $g_i\sim\mathrm{Gumbel}(0,1)$.
Define the unperturbed baseline
\begin{equation}
\mathbf{p}_{\tau} \;\triangleq\; \sigma\!\left(\tau^{-1}\log\boldsymbol{\alpha}\right).
\end{equation}
We analyze the expected loss in a neighborhood of $\mathbf{p}_{\tau}$ under the small-perturbation regime (defined below). 
Then, for a twice-differentiable loss $\mathcal{L}(\theta,\mathbf{p})$ that depends on $\mathbf{p}$ through the forward computation (hence generally non-linear in $\mathbf{p}$), we have:
\begin{equation}
\mathbb{E}_{\mathbf{g}}\!\left[\mathcal{L}(\theta,\tilde{\mathbf{p}})\right]
=
\mathcal{L}(\theta,\mathbf{p}_{\tau})
+
\frac{\pi^2}{12\tau^2}\,
\mathrm{Tr}\!\Big(
\mathbf{J}_{\sigma}(\mathbf{p}_{\tau})^\top
\,\nabla_{\mathbf{p}}^2\mathcal{L}(\theta,\mathbf{p})\big|_{\mathbf{p}=\mathbf{p}_{\tau}}\,
\mathbf{J}_{\sigma}(\mathbf{p}_{\tau})
\Big)
+
\mathcal{O}\!\Big(\|\nabla_{\mathbf{p}}\mathcal{L}\|\,\mathbb{E}\|\boldsymbol{\delta}\|^2\Big)
+
\mathcal{O}\!\Big(\mathbb{E}\|\boldsymbol{\delta}\|^3\Big),
\label{eq:theorem1_final}
\end{equation}
\textit{where $\mathbf{J}_{\sigma}(\mathbf{p}_{\tau})=\mathrm{diag}(\mathbf{p}_{\tau})-\mathbf{p}_{\tau}\mathbf{p}_{\tau}^\top$ is the Softmax Jacobian evaluated at $\mathbf{p}_{\tau}$, and $\boldsymbol{\delta}$ is the effective (zero-mean) perturbation in logit space defined in the proof. In particular, in our experiments with $\tau=1$, $\mathbf{p}_{\tau}=\sigma(\log\boldsymbol{\alpha})=\boldsymbol{\alpha}$, so the expansion is centered at the semantic target. This indicates that Gumbel noise induces a curvature penalty that favors flatter neighborhoods around the target distribution, with leading-order scaling $1/\tau^2$.}
}

\textit{Proof.}

\textbf{1. Notation and mean removal via translation invariance.}
Let $\mathbf{l}=\log\boldsymbol{\alpha}$.
The stochastic target is
\begin{equation}
\tilde{\mathbf{p}}
=
\sigma\!\left(\frac{\mathbf{l}+\mathbf{g}}{\tau}\right).
\end{equation}
A standard Gumbel variable has mean $\mathbb{E}[g]=\gamma$ and variance $\mathrm{Var}(g)=\pi^2/6$.
Decompose $\mathbf{g}=\boldsymbol{\epsilon}+\gamma\mathbf{1}$ where $\mathbb{E}[\boldsymbol{\epsilon}]=\mathbf{0}$ and
$\mathbb{E}[\boldsymbol{\epsilon}\boldsymbol{\epsilon}^\top]=\frac{\pi^2}{6}\mathbf{I}$.
Using Softmax translation invariance $\sigma(\mathbf{x}+c\mathbf{1})=\sigma(\mathbf{x})$, we obtain
\begin{equation}
\tilde{\mathbf{p}}
=
\sigma\!\left(\frac{\mathbf{l}+\boldsymbol{\epsilon}}{\tau}\right)
=
\sigma\!\left(\frac{\mathbf{l}}{\tau}+\boldsymbol{\delta}\right),
\qquad
\boldsymbol{\delta}\triangleq \frac{\boldsymbol{\epsilon}}{\tau}.
\end{equation}
Define the baseline $\mathbf{p}_{\tau}\triangleq\sigma(\mathbf{l}/\tau)$, so that $\tilde{\mathbf{p}}=\sigma(\mathbf{z}_\tau+\boldsymbol{\delta})$ with $\mathbf{z}_\tau=\mathbf{l}/\tau$ and $\tilde{\mathbf{p}}(\mathbf{0})=\mathbf{p}_\tau$.

\textbf{2. Second-order expansions.}
We assume a small-perturbation regime where $\|\boldsymbol{\delta}\|$ is sufficiently small such that second-order Taylor expansions are valid and the remainders are controlled by $\mathcal{O}(\|\boldsymbol{\delta}\|^3)$ in expectation.

\emph{Softmax expansion.}
Let $\mathcal{H}_{\sigma}$ denote the (third-order) Hessian tensor of Softmax with respect to logits.
Using tensor contraction notation $\mathcal{H}_{\sigma}[\boldsymbol{\delta},\boldsymbol{\delta}]\in\mathbb{R}^V$,
\begin{equation}
\tilde{\mathbf{p}}-\mathbf{p}_{\tau}
=
\mathbf{J}_{\sigma}(\mathbf{p}_{\tau})\,\boldsymbol{\delta}
+
\frac{1}{2}\,\mathcal{H}_{\sigma}[\boldsymbol{\delta},\boldsymbol{\delta}]
+
\mathcal{O}(\|\boldsymbol{\delta}\|^3).
\end{equation}

\emph{Loss expansion.}
Expanding $\mathcal{L}(\theta,\mathbf{p})$ around $\mathbf{p}=\mathbf{p}_{\tau}$,
\begin{align}
\mathcal{L}(\theta,\tilde{\mathbf{p}})
&=
\mathcal{L}(\theta,\mathbf{p}_{\tau})
+
(\tilde{\mathbf{p}}-\mathbf{p}_{\tau})^\top
\nabla_{\mathbf{p}}\mathcal{L}(\theta,\mathbf{p})\big|_{\mathbf{p}=\mathbf{p}_{\tau}}
+
\frac{1}{2}
(\tilde{\mathbf{p}}-\mathbf{p}_{\tau})^\top
\nabla_{\mathbf{p}}^2\mathcal{L}(\theta,\mathbf{p})\big|_{\mathbf{p}=\mathbf{p}_{\tau}}
(\tilde{\mathbf{p}}-\mathbf{p}_{\tau})
+
\mathcal{O}(\|\tilde{\mathbf{p}}-\mathbf{p}_{\tau}\|^3).
\end{align}
Substituting the first-order relation $\tilde{\mathbf{p}}-\mathbf{p}_{\tau}=\mathbf{J}_{\sigma}\boldsymbol{\delta}+\mathcal{O}(\|\boldsymbol{\delta}\|^2)$ into the quadratic term yields
\begin{equation}
\mathcal{L}(\theta,\tilde{\mathbf{p}})
=
\mathcal{L}(\theta,\mathbf{p}_{\tau})
+
\left(\mathbf{J}_{\sigma}\boldsymbol{\delta}+\frac12\mathcal{H}_{\sigma}[\boldsymbol{\delta},\boldsymbol{\delta}]\right)^\top
\nabla_{\mathbf{p}}\mathcal{L}\big|_{\mathbf{p}=\mathbf{p}_{\tau}}
+
\frac{1}{2}\boldsymbol{\delta}^\top
\mathbf{J}_{\sigma}^\top
\nabla_{\mathbf{p}}^2\mathcal{L}\big|_{\mathbf{p}=\mathbf{p}_{\tau}}
\mathbf{J}_{\sigma}\,
\boldsymbol{\delta}
+
\mathcal{O}(\|\boldsymbol{\delta}\|^3),
\end{equation}
where $\mathbf{J}_\sigma=\mathbf{J}_{\sigma}(\mathbf{p}_\tau)$ for brevity.

\textbf{3. Taking expectations.}
Since $\mathbb{E}[\boldsymbol{\delta}]=\mathbf{0}$, the linear-in-$\boldsymbol{\delta}$ term vanishes:
$\mathbb{E}[\mathbf{J}_{\sigma}\boldsymbol{\delta}]=\mathbf{0}$.
The remaining first-order bias arises from Softmax nonlinearity:
\begin{equation}
\mathbb{E}\!\left[\left(\tfrac12\mathcal{H}_{\sigma}[\boldsymbol{\delta},\boldsymbol{\delta}]\right)^\top
\nabla_{\mathbf{p}}\mathcal{L}\big|_{\mathbf{p}=\mathbf{p}_{\tau}}
\right]
=
\mathcal{O}\!\Big(\|\nabla_{\mathbf{p}}\mathcal{L}\|\,\mathbb{E}\|\boldsymbol{\delta}\|^2\Big),
\end{equation}
which becomes negligible near a stationary point where $\|\nabla_{\mathbf{p}}\mathcal{L}\|\approx 0$.

For the quadratic term,
\begin{align}
\mathbb{E}\!\left[\frac12\boldsymbol{\delta}^\top
\mathbf{J}_{\sigma}^\top
\nabla_{\mathbf{p}}^2\mathcal{L}\big|_{\mathbf{p}=\mathbf{p}_{\tau}}
\mathbf{J}_{\sigma}\,
\boldsymbol{\delta}\right]
&=
\frac12\,\mathrm{Tr}\!\left(
\mathbf{J}_{\sigma}^\top
\nabla_{\mathbf{p}}^2\mathcal{L}\big|_{\mathbf{p}=\mathbf{p}_{\tau}}
\mathbf{J}_{\sigma}\,
\mathbb{E}[\boldsymbol{\delta}\boldsymbol{\delta}^\top]
\right).
\end{align}
Using $\mathbb{E}[\boldsymbol{\epsilon}\boldsymbol{\epsilon}^\top]=\frac{\pi^2}{6}\mathbf{I}$ and $\boldsymbol{\delta}=\boldsymbol{\epsilon}/\tau$ gives
$\mathbb{E}[\boldsymbol{\delta}\boldsymbol{\delta}^\top]=\frac{\pi^2}{6\tau^2}\mathbf{I}$,
hence
\begin{equation}
\mathbb{E}\!\left[\frac12\boldsymbol{\delta}^\top
\mathbf{J}_{\sigma}^\top
\nabla_{\mathbf{p}}^2\mathcal{L}\big|_{\mathbf{p}=\mathbf{p}_{\tau}}
\mathbf{J}_{\sigma}\,
\boldsymbol{\delta}\right]
=
\frac{\pi^2}{12\tau^2}\,
\mathrm{Tr}\!\Big(
\mathbf{J}_{\sigma}^\top
\nabla_{\mathbf{p}}^2\mathcal{L}\big|_{\mathbf{p}=\mathbf{p}_{\tau}}
\mathbf{J}_{\sigma}
\Big).
\end{equation}
Collecting terms yields Eq.~(\ref{eq:theorem1_final}), with the remainder $\mathcal{O}(\mathbb{E}\|\boldsymbol{\delta}\|^3)$ from Taylor truncation. \hfill $\square$




\section{Construction of the Multi-Chain GSM8k-Aug Dataset}
\label{app_dataset}
\begin{figure*}[t]
  \centering
  \includegraphics[width=\linewidth]{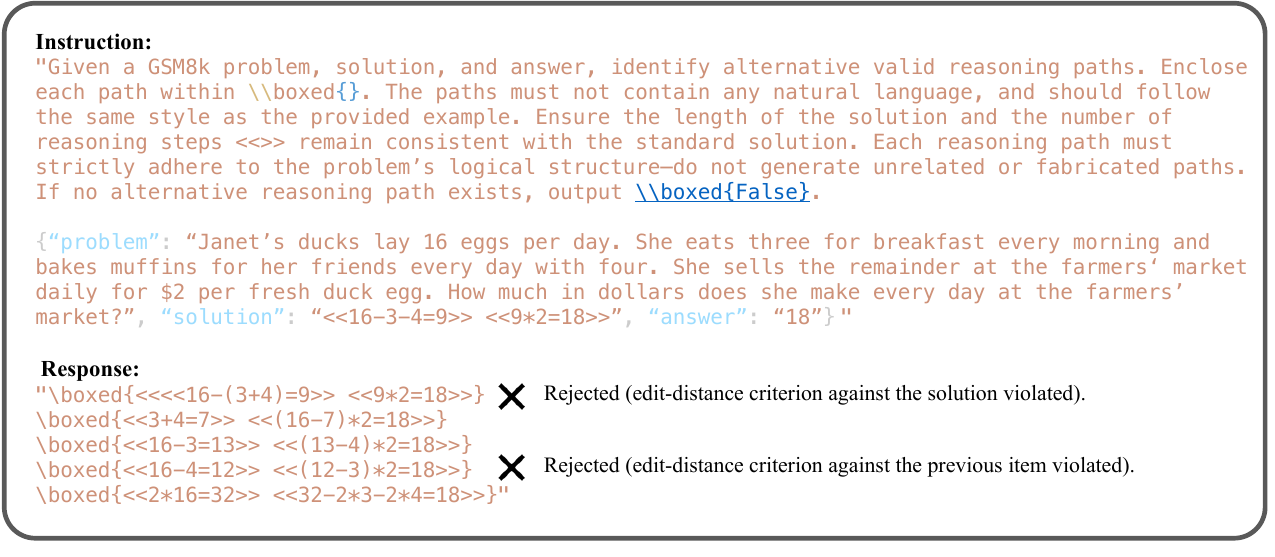}
  \caption{Illustration of the Multi-Chain dataset construction pipeline. We prompt GPT-5 to generate alternative valid reasoning paths for a given GSM8k problem. The response section demonstrates the diversity filtering mechanism: candidate paths are evaluated based on edit distance, and those exhibiting high similarity to the standard solution or previously accepted paths (marked with $\times$) are rejected to ensure significant reasoning diversity.}
  \label{fig_dataset_case}
\end{figure*}

We carefully designed prompts and employed the advanced GPT-5 model to generate preliminary candidate reasoning chains. Given the simplicity and limited number of questions, we manually inspected and removed examples that either failed to follow the reasoning format or contained incorrect logic. To further maximize diversity, we retained only reasoning chains with pairwise similarity below 0.9 based on edit distance. An example prompt call is illustrated in Figure~\ref{fig_dataset_case}. Among the 1,319 data points, 1,310 contained multiple reasoning paths. 

\begin{figure*}[t]
  \centering
  \includegraphics[width=\linewidth]{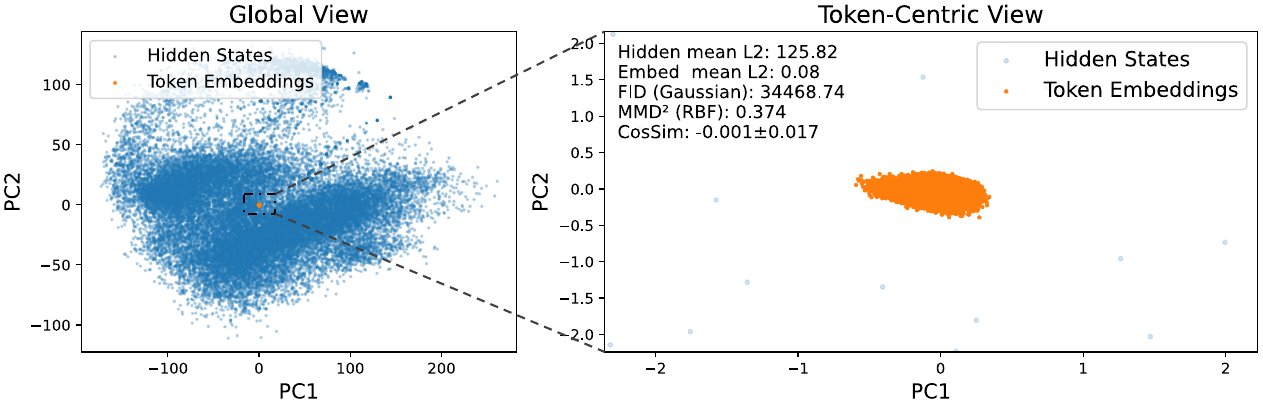}
  \caption{Visualization of last-layer hidden states and token embeddings of the Deepseek-Distill-Qwen-7B model on the Math500 dataset.}
  \label{fig_hidden_vs_embeddings_deepseek}
\end{figure*}

\section{Analytical Metrics}
\subsection{Notation and Preliminaries}
Let $\mathcal{V}$ be the vocabulary, $|\mathcal{V}|=V$. For a given problem, fix a reference explicit chain (from an explicit CoT solution)
\begin{equation}
\bm{x} = (x_1,\ldots,x_{L_{\text{exp}}}),\quad x_i \in \mathcal{V}.
\end{equation}
A latent model generates a latent reasoning chain consisting of $T$ latent steps, each with a vocabulary distribution
\begin{equation}
p_t \in \Delta^{V-1}\quad (t=1,\ldots,T),
\end{equation} 
where $\Delta^{V-1}$ is the probability simplex. To quantify the extent of explicit information captured by each latent step (whether from a single chain or multiple chains), we introduce:

\textbf{Assumption 1.} During training, if each latent token is constrained to receive information from a fixed number of explicit tokens, then at inference time the amount of explicit information (in terms of token granularity) captured by each latent step is uniform.

This assumption is reasonable, as the model is expected to behave consistently during training and inference. So we align the two sequences by a ratio $r\in \mathbb{N}$: every $r$ explicit tokens are grouped to one latent step. Define the number of aligned steps
\begin{equation}
T' = \min\bigl(T, \lceil L_{\exp}/r \rceil\bigr),
\end{equation} 
and the explicit token set aligned to latent step $t$:
\begin{equation}
S_t=\left\{x_{(t-1)r+1},\,\ldots,\,x_{\min(tr,L_{\exp})}\right\},\qquad |S_t|\le r.
\end{equation}

Let $\mathcal{T}_{K}(p_t) \subset \mathcal{V}$ be the set of the $K$ highest-probability tokens under $p_t$.

\subsection{ECR@K (Effective Compression Rate)}
\label{app_ecr}
To verify whether latent reasoning corresponds to the compression of a single chain, we examine, \textit{on average, how many aligned explicit tokens from the reference chain appear within the Top-K probabilities at each latent step}. Thus, we define
\begin{equation}
\mathrm{ECR@K} \;=\; \frac{1}{T'} \sum_{t=1}^{T'} \bigl| S_t \cap \mathcal{T}_K(p_t) \bigr| \;\in\; [0,r].
\end{equation}

A value greater than 1 indicates genuine compression of a single reasoning chain, meaning that each latent step, on average, covers more than one token from the reference explicit chain.
\subsection{N$_{\mathrm{eff}}$ and Top-2 Score}
\label{app_neff}
Given $M$ candidate explicit chains $\{\bm{x}^{(m)}\}_{m=1}^M$ (with lengths $L_m$), we align each chain to the $T$ latent steps using the same ratio $r$ as above. Let the explicit tokens aligned to step $t$ on chain $m$ be $S_{m,t}\subseteq\mathcal{V}$ (at most $r$ tokens), and let $\mathcal{T}_{K}(p_t)$ be the Top-$K$ tokens under $p_t$. Define the per-step \textit{mass} of chain $m$ as
\begin{equation}
\mathrm{mass}_{m,t}^{(K)} \;=\; \sum_{x\in S_{m,t}\cap \mathcal{T}_{K}(p_t)} p_t[x],
\end{equation}
and aggregate it across steps to obtain a sequence-level evidence score
\begin{equation}
\mathrm{score}_m \;=\; \frac{1}{T'} \sum_{t=1}^{T'} \log\!\big(\mathrm{mass}_{m,t}^{(K)} + \varepsilon\big),
\end{equation}
where $T'=\min\!\big(T,\lceil L_m/r\rceil\big)$, $\varepsilon>0$ is a small constant, and $K$ controls robustness (setting $K=V$ recovers the full-mass version).
We convert $\{\mathrm{score}_m\}$ to a posterior over paths via a temperatured softmax
\begin{equation}
P_m \;=\; \frac{\exp(\mathrm{score}_m/\tau)}{\sum_{j=1}^M \exp(\mathrm{score}_j/\tau)} \quad (\tau=1).
\end{equation}

The \textbf{effective number of supported paths} is
\begin{equation}
\mathrm{N}_{\mathrm{eff}} \;=\; \exp\!\Big(-\sum_{m=1}^M P_m \log P_m\Big),
\end{equation}
and the \textbf{Top-2 score} is the ratio between the two largest posteriors,
\begin{equation}
\mathrm{Top\mbox{-}2} \;=\; \frac{P_{(2)}}{P_{(1)}} \quad\text{with}\quad P_{(1)}\ge P_{(2)}.
\end{equation}
Intuitively, a larger $\mathrm{score}_m$ means the model consistently places high probability on chain $m$ along the sequence; a larger N$_{\mathrm{eff}}$ and higher Top-2 indicate stronger parallel support across multiple chains rather than collapse to a single path.

When only a single path exists in parallel, N$_{\mathrm{eff}}=1$. However, since different paths may partially overlap in their prefixes, we adopt a conservative baseline of N$_{\mathrm{eff}}=1.7$ (corresponding to approximately 70\% probability mass on path 1 and 30\% on path 2). Values exceeding this threshold are taken as evidence of parallel inference. In addition, to mitigate the influence of tail noise, we set $K=100$.

\begin{figure*}[t]
  \centering
  \includegraphics[width=\linewidth]{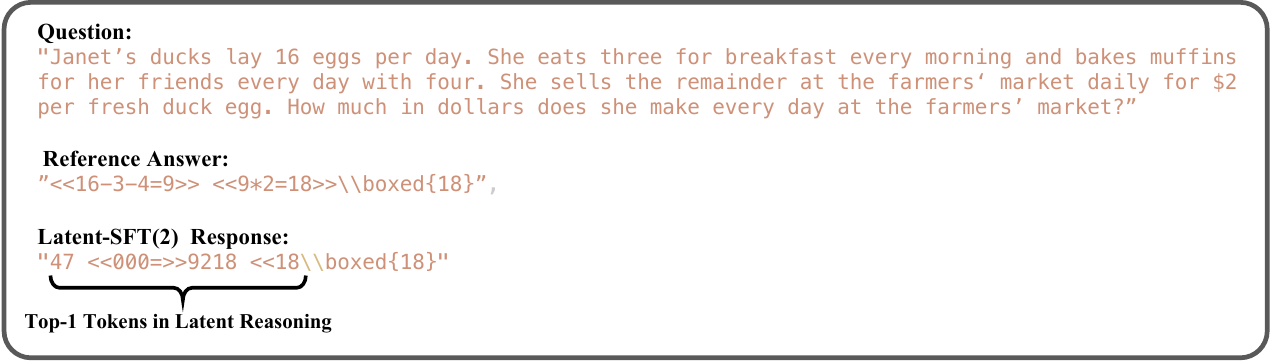}
  \caption{Reasoning example of Latent-SFT(2) on the GSM8k-Aug test set. The model generates a sequence of latent tokens (annotated as "Top-1 Tokens") instead of explicit natural language steps. While these tokens appear as unintelligible numeric combinations to humans, they effectively encode the necessary logical computations, allowing the model to derive the correct answer 18. Notably, this latent process reduces the reasoning chain length by nearly 50\% compared to the reference solution, demonstrating high computational efficiency.}
  \label{fig_gsm8k_case}
\end{figure*}

\end{document}